\documentclass[letterpaper, 10 pt, conference]{ieeeconf}  

\IEEEoverridecommandlockouts                              

\usepackage{graphics} 
\usepackage{amsmath} 
\usepackage{amssymb}  
\usepackage{algorithm}
\usepackage{algorithmic}
\usepackage{xcolor}
\usepackage{graphicx}
\usepackage{subcaption}
\usepackage{balance}
\usepackage{multirow}
\usepackage{booktabs}
\usepackage[top=20.1mm, bottom=15.2mm, left=16.9mm, right=16.9mm]{geometry}
\usepackage{cite}
\makeatletter
\let\NAT@parse\undefined
\makeatother
\usepackage[colorlinks,linkcolor=blue,citecolor=blue]{hyperref}

\title{\LARGE \bf
MA-ROESL: Motion-aware Rapid Reward Optimization for Efficient Robot Skill Learning from Single Videos
}

\author{Xianghui Wang$^{1*}$, Xinming Zhang$^{1, 2*}$, Yanjun Chen$^{3}$, Xiaoyu Shen$^{1}$ and Wei Zhang$^{1,\dagger}$
\thanks{This work has been submitted to the IEEE for possible publication. Copyright may be transferred without notice, after which this version may no longer be accessible.}%
\thanks{*This work is supported by 2035 Key Research and Development Program of Ningbo City under Grant No.2024Z127.}
\thanks{$^{1}$The authors are with the College of Information Science and Technology, Eastern Institute of Technology, Ningbo, China. {$^{*}$ are equal contributors, $^{\dagger}$ is the corresponding author. } {\tt\small \{xhwang, xmzhang, xyshen, zhw\}@eitech.edu.cn}}%
\thanks{$^{2}$The author is with the School of Computer Science and Technology, The University of Science and Technology of China, China. {\tt\small  xm\_zhang@mail.ustc.edu.cn}}%
\thanks{$^{3}$The author is with the Department of Computing, The Hong Kong Polytechnic University, Hong Kong SAR, China. {\tt\small yan-jun.chen@connect.polyu.hk}}
}

\begin{document}
\maketitle
\thispagestyle{empty}
\pagestyle{empty}
\begin{abstract}
Vision-language models (VLMs) have demonstrated excellent high-level planning capabilities, enabling locomotion skill learning from video demonstrations without the need for meticulous human-level reward design. However, the improper frame sampling method and low training efficiency of current methods remain a critical bottleneck, resulting in substantial computational overhead and time costs. To address this limitation, we propose Motion-aware Rapid Reward Optimization for Efficient Robot Skill Learning from Single Videos (MA-ROESL). MA-ROESL integrates a motion-aware frame selection method to implicitly enhance the quality of VLM-generated reward functions. It further employs a hybrid three-phase training pipeline that improves training efficiency via rapid reward optimization and derives the final policy through online fine-tuning. Experimental results demonstrate that MA-ROESL significantly enhances training efficiency while faithfully reproducing locomotion skills in both simulated and real-world settings, thereby underscoring its potential as a robust and scalable framework for efficient robot locomotion skill learning from video demonstrations.
\end{abstract}

\section{INTRODUCTION}
Deep Reinforcement Learning (DRL) has achieved remarkable progress in robot locomotion control, enabling robots to autonomously learn locomotion skills~\cite{rudin2022learning,vollenweider2023advanced,margolis2023walk}. However, these methods often require meticulous hand-designed reward functions and a labor-intensive tuning process~\cite{escontrela2022adversarial}. This shortage becomes even more pronounced when developing highly dynamic locomotion skills \cite{li2023learning}.

Recent research has increasingly focused on learning skills directly from real-world videos \cite{sontakke2023roboclip, li2024sds}. This approach leverages advances in VLMs, which can extract high-level semantic representations in video demonstrations and translate them into reward functions. Although motion capture (Mocap) data has proven effective for training simulated controllers, it requires highly instrumented environments and human actors, making large-scale collection difficult and costly \cite{wagener2022mocapact}. In contrast, online videos offer a rich and diverse source of locomotion skill demonstrations, encompassing a wide range of species and environments. This diversity is reflected in the fact that over 300 hours of video are uploaded to YouTube every minute \cite{peng2018sfv}. This abundance, together with the rich semantic information embedded in videos, makes videos a promising alternative for learning diverse and naturalistic quadruped locomotion skills, especially in scenarios where collecting high-quality Mocap data is infeasible.

Despite its growing promise, learning locomotion skills from single videos faces two key challenges. \textbf{The first challenge arises from improper frame sampling methods that fail to accurately capture motion patterns in videos.} Existing methods typically adopt uniform frame sampling, which disregards the temporal structure of periodic patterns~\cite{li2024sds}. For instance, quadruped locomotion relies on cyclic limb coordination, and omitting critical transitions between the stance and swing phases can mislead the generation of rewards. \textbf{The second challenge lies in the inability to evaluate the effectiveness of VLM-generated rewards prior to full policy training.} When the reward fails to foster effective skill learning, significant computational overhead is incurred due to unnecessary training cycles, ultimately degrading overall training efficiency.

To address this limitation, we introduce Motion-aware Rapid Reward Optimization for Efficient Robot Skill Learning from Single Videos (MA-ROESL), a framework that enables efficient skill learning from video demonstrations. It employs a motion-aware frame selection method that focuses on behaviorally salient frames to implicitly enhance the quality of VLM-generated reward functions. In addition, it adopts a hybrid three-phase training pipeline that enables efficient skill learning through rapid offline reward optimization followed by online policy fine-tuning. Extensive experiments demonstrate that MA-ROESL significantly improves training efficiency. Furthermore, in both simulated and real-world settings, the robot reliably reproduces the target skills from video demonstrations, validating MA-ROESL’s capability to effectively learn locomotion skills from video demonstrations and achieve sim-to-real zero-shot transfer. The key contributions of this work are summarized as follows:

\begin{enumerate}
    \item A motion-aware frame selection method is proposed to better reflect video demonstrations and implicitly improve VLM-generated reward functions.
    \item A hybrid three-phase training pipeline is proposed to improve the training efficiency of skill learning from video demonstrations.  
    \item MA-ROESL is extensively evaluated on diverse quadruped locomotion skills, demonstrating improved training efficiency and effective zero-shot sim-to-real transfer.
\end{enumerate}

\section{RELATED WORKS}
\subsection{Learning from Video Demonstration}
In DRL, the reward function serves as a core component that defines the objective and guides the agent’s learning process \cite{liu2021deep,zhang2024achieving}. For tasks with explicit rules \cite{silver2017mastering,zhang2025drl,goldwaser2020deep}, the design of rewards is relatively straightforward and interpretable. However, it remains highly challenging for problems with long horizons and high-dimensional state spaces \cite{elguea2025novel}. Consequently, learning from video directly has received increasing attention, with methods relying on task-independent rewards or visual representations \cite{zhou2018deep,liu2018imitation,peng2018sfv,schmeckpeper2020reinforcement,DBLP:conf/rss/PengCZLTL20}. Nevertheless, these methods frequently suffer from limited performance and high labor costs.

In recent years, LLMs have demonstrated outstanding performance as high-level semantic planners in robotics \cite{singh2023progprompt}. Several studies have explored using LLMs to directly generate reward function code for guiding robot learning. For example, \textit{Yu et al.} investigated LLM-assisted reward design, but required domain-specific descriptions and predefined reward templates \cite{yu2023language}. Further, EUREKA explored the use of LLMs to autonomously generate reward functions without human intervention, combined with evolutionary search to identify effective reward strategies \cite{ma2024eureka}. DrEureka extended this line of work by automating the entire sim-to-real procedure, including the design of both reward functions and domain randomization configurations \cite{ma2024dreureka}.

However, when handling complex tasks, natural language often faces limitations in expressing the task complexity. To address this, RoboCLIP \cite{sontakke2023roboclip} integrates VLMs with RL by encoding both visual and language information into a shared embedding space. Similarly, \cite{rocamonde2023vision} employs a VLM as a reward model to guide agent learning. Closest to our work is \cite{li2024sds}, which explores using VLMs to automatically generate, evaluate, and improve reward functions, thereby enabling skill learning from a single video without manual annotation. However, \cite{li2024sds} relies on online training, which requires substantial time for reward evaluation and optimization. This issue also appears in \cite{sontakke2023roboclip, zhang2023slomo}, where the iterative evaluation of reward functions is extremely time-consuming and computationally inefficient. This significant bottleneck severely limits the feasibility of scaling to a wide range of tasks and poses a major challenge for broad real-world deployment.

\subsection{Offline Reinforcement Learning}
While online RL enables strong exploration and real-time adaptability through agent-environment interaction \cite{DBLP:journals/corr/LillicrapHPHETS15}, it suffers from low sample efficiency and high computational cost \cite{prudencio2023survey,kiran2021deep}. Instead, offline RL, which learns entirely from a fixed dataset of previously collected trajectories without agent-environment interaction, offers unprecedented advantages in training efficiency \cite{levine2020offline}. However, the mismatch between the behavior policy and the target policy leads to distributional shift, potentially resulting in value overestimation \cite{prudencio2023survey}. This challenge is amplified in robotic control, where inaccurate value estimates can result in unsafe actions and severe deployment failures.

To address this, model-free methods typically adopt one of two strategies.
The first is to restrict policy updates to remain close to the behavior policy. For instance, Batch-Constrained Q-Learning (BCQ) employs a generative model to limit action selection to those likely under the dataset’s behavior distribution \cite{pmlr-v97-fujimoto19a}. The second is to penalize overestimated Q-values for out-of-distribution actions. Conservative Q-Learning (CQL) achieves this by adding a regularization term that lowers Q-values for unseen state-action pairs, producing more conservative estimates \cite{kumar2020conservative}. Furthermore, Implicit Q-Learning (IQL) estimates an upper bound of the state value function via expectile regression over dataset actions, thereby avoiding explicit queries on out-of-distribution actions during value updates and enabling stable policy improvement. Despite recent progress in offline RL, the lack of real-time agent-environment interaction often leads to suboptimal policies and limited generalization in dynamic environments.

\section{Problem Formulation}
\begin{figure}
    \centering
    \includegraphics[width=0.9\linewidth]{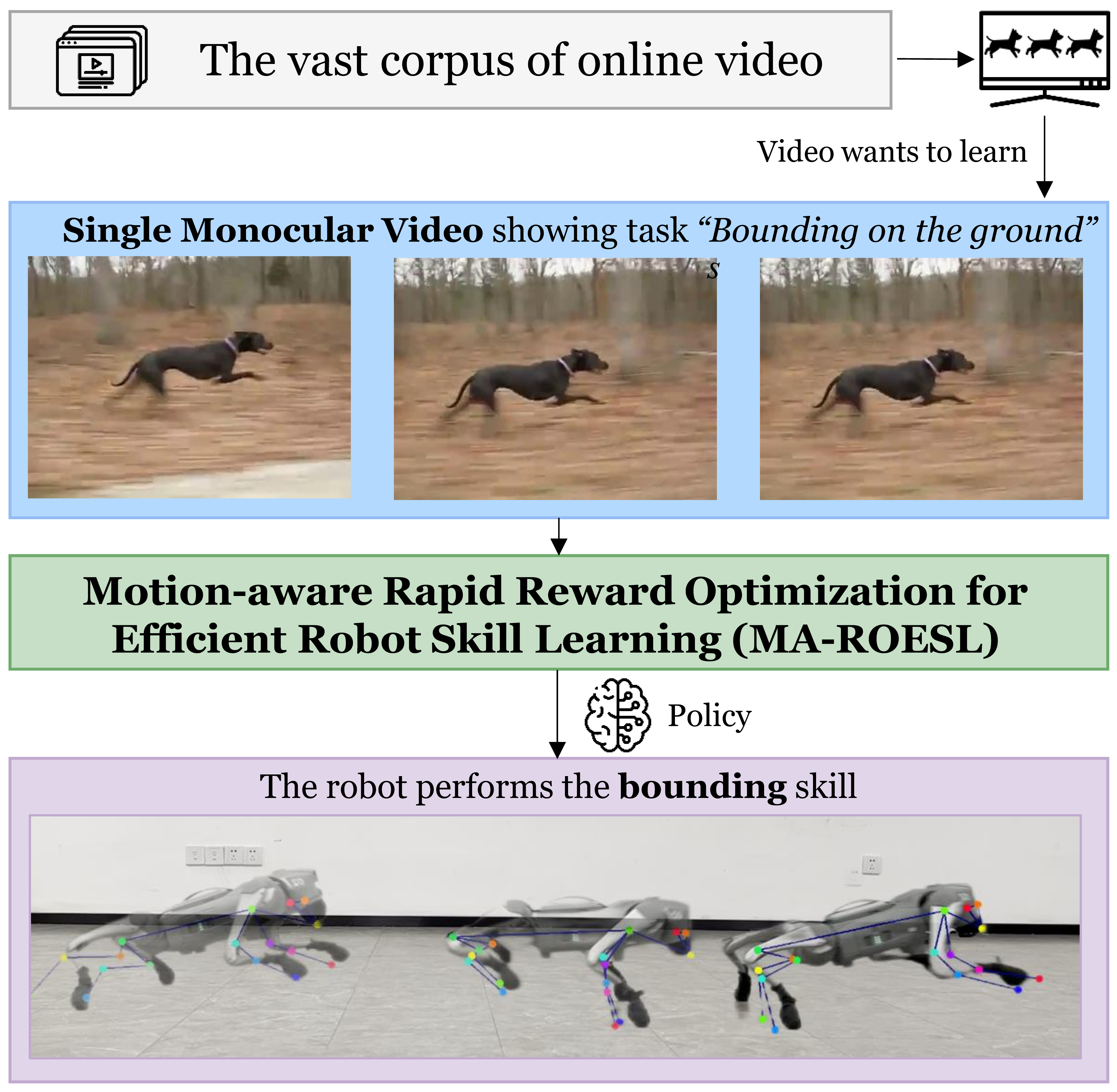}
    \caption{Illustration of the skill learning problem from video demonstrations.}
    \label{fig:problem_formulation}
\end{figure}

This section formally defines the problem of learning locomotion skills from video demonstrations, which is illustrated in Fig.~\ref{fig:problem_formulation}. This process can be formulated as a Reward Design Problem (RDP)~\cite{singh2009rewards}, represented as a tuple $P = \langle M, \mathcal{R}, \pi_M, F \rangle$, where $M = (\mathcal{S}, \mathcal{A}, \mathcal{T})$, $\mathcal{S}$ denotes the state space, $\mathcal{A}$ denotes the action space, and $\mathcal{T}$ denotes the transition function~\cite{ma2023eureka}. In addition, a VLM is used for reward design, which takes the world model $M$ and a video demonstration $v$ as inputs and outputs a set of candidate reward functions $\mathcal{R} = \{R_1, R_2, \dots, R_k\}$, as depicted in (\ref{eq:reward_set}):

\begin{equation}
    \mathcal{R} = \texttt{VLM}(M, v)
    \label{eq:reward_set}
\end{equation}

Assume a policy learning algorithm $A(\cdot)$ for a specific Markov Decision Process (MDP), which outputs a policy $\pi$. The policy $\pi$ optimizes the reward $R \in \mathcal{R}$ within the MDP $(M,R)$, as represented in (\ref{eq:pi_A(M,R)}).

\begin{equation}
    \pi \leftarrow A(M, R)
    \label{eq:pi_A(M,R)}
\end{equation}

For each skill, the fitness function is defined as $F:\Pi \to \mathbb{R}$, mapping a policy to a scalar value that reflects the policy's performance within the unknown true MDP $M_{\text{un}}$, as shown in (\ref{eq:f=F(pi)}).

\begin{equation}
    f = F_{M_{\text{un}}}(\pi)
    \label{eq:f=F(pi)}
\end{equation}

In the RDP, the objective is to output a reward function $R \in \mathcal{R}$ that maximizes the fitness function $F$, which can be expressed in (\ref{eq:maximizef}).

\begin{equation}
    \max_{R \in \mathcal{R}} F_{M_\text{un}}(A(M,R))
    \label{eq:maximizef}
\end{equation}

One key challenge lies in selecting the best reward function \( R \in \mathcal{R} \), which demands extensive training iterations. This process is extremely time-consuming~\cite{li2024sds}, inducing inefficiency that ultimately impairs scalability.

\section{METHODS}
This section introduces the core components of MA-ROESL and their roles in addressing the underlying optimization objective (\ref{eq:maximizef}). The framework comprises two functionally complementary modules: a motion-aware frame selection method (Section \ref{method:motion_aware_frame_selection}) and a hybrid three-phase training pipeline (Section \ref{method:MA-ROESL}). The motion-aware frame selection method improves the fidelity of video representations by selecting behaviorally salient frames, thereby implicitly enhancing the quality of VLM-generated reward functions. Building upon these enhanced inputs, the hybrid three-phase training pipeline improves training efficiency by rapidly optimizing reward functions and subsequently fine-tuning the final policy. This compositional design enables MA-ROESL to efficiently identify the best reward function $R$ that maximizes policy performance $F$ in the unknown true MDP $M_\text{un}$.

\subsection{Motion-aware Frame Selection}
\label{method:motion_aware_frame_selection}
Similar to~\cite{li2024sds}, the input to the VLM consists of frames sampled from the demonstration video. However, the uniform sampling method treats all frames equally and often neglects temporal dynamics, resulting in redundant or uninformative selections. To address this limitation, a motion-aware frame selection method is proposed to prioritize temporally dynamic segments. Specifically, as shown in (\ref{eq:dense_optical_flow}), motion-salient frames are identified by analyzing the dense optical flow between consecutive frames~\cite{farneback2003two}.

\begin{equation}
\sigma_k = \frac{1}{W \times H} \sum_{i=1}^{W} \sum_{j=1}^{H} \left\| \mathbf{d}_k(i, j) \right\|_2
\label{eq:dense_optical_flow}
\end{equation}
where $\sigma_k$ represents the average motion scores between frame $k-1$ and frame $k$. $W$ and $H$ denote the width and height of the frame, respectively. $\mathbf{d}_k(i, j)$ denotes the dense optical flow displacement at pixel $(i, j)$. Based on the computed motion scores, frames corresponding to the most pronounced movements are selected. To maintain sufficient temporal coverage and avoid bias toward short-duration motions, additional frames are uniformly sampled if the initial selection is insufficient. The set of selected frames, denoted as $\mathcal{K}{\text{motion}}$, is defined in (\ref{eq:keyframe_seq}).

\begin{equation}
\mathcal{K}_{\text{motion}} = \left\{ k \in \{1, \dots, T-1\} \mid \text{Rank}(\sigma_k) \le K \right\}
\label{eq:keyframe_seq}
\end{equation}
where $T$ is the total number of frames, $\text{Rank}(\sigma_k)$ denotes the descending order of $\sigma_k$, and \( K \) is the target number of motion-salient frames to be selected.

\begin{figure}[t]
    \centering    \includegraphics[width=0.8\linewidth]{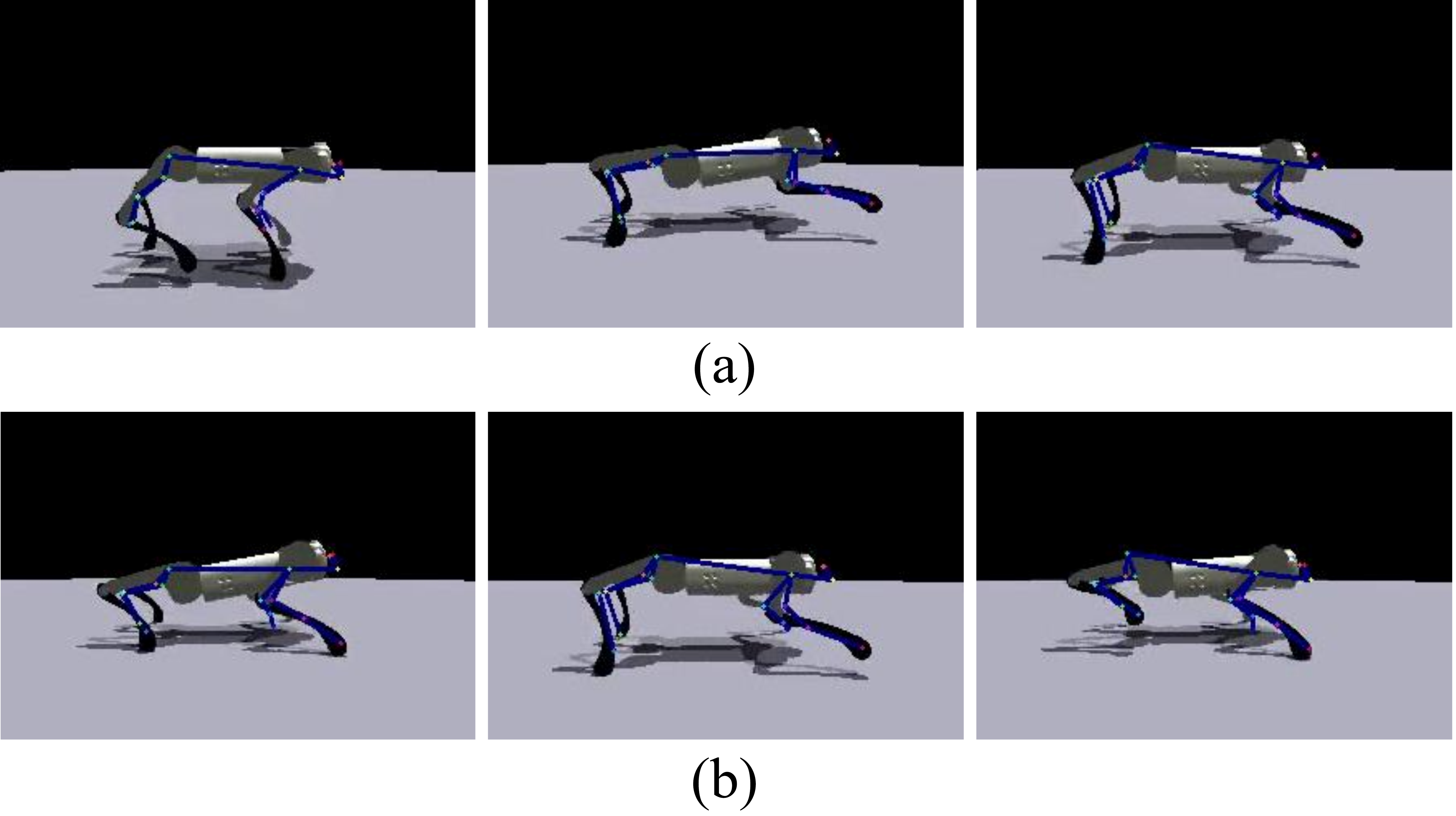}
    \caption{The comparison of frame selection results: a) uniform sampling method. b) motion-aware frame selection method.}
    \label{fig:sample-method-comparison}
\end{figure}

Fig.~\ref{fig:sample-method-comparison} compares two frame selection methods. As shown, in the bound skill, the robot's front and rear legs alternate ground contacts. However, the uniform sampling method selects frames when the front legs are lifted, neglecting the corresponding rear leg actions. This bias compromises the representational accuracy of the visual input and may mislead the VLM in interpreting the robot's motion patterns. To address this issue, the motion-aware frame selection method targets temporally salient motion segments, ensuring that the selected frames accurately reflect the robot's actual movements. This method preserves movement information and provides the VLM with more accurate and informative visual cues for reward function generation.

\begin{figure*}[t]
    \centering
    \includegraphics[width=0.90\textwidth]{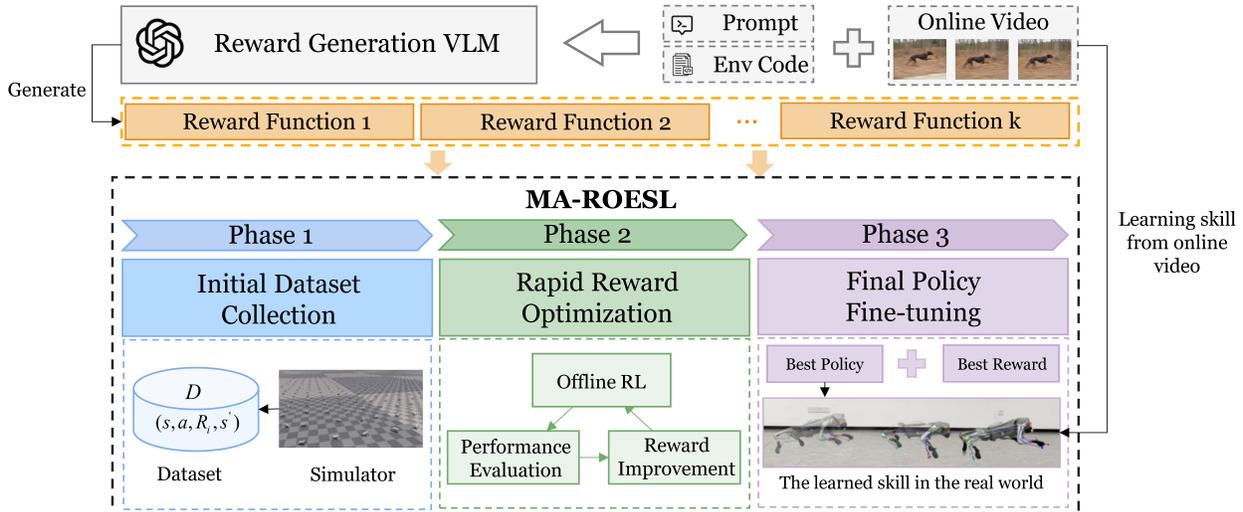}
    \caption{Overview of MA-ROESL framework. 1) Phase 1: VLM-generated reward functions are used to train policies and collect an offline dataset for subsequent reward optimization. 2) Phase 2: Offline RL enables rapid reward optimization by relabeling rewards using the dataset and selecting the best-performing reward-policy pair. 3) Phase 3: The selected policy is fine-tuned online under the same reward to improve robustness in real-world deployment.}
    \label{fig:sample}
\end{figure*}

\subsection{Hybrid Three-phase Training Pipeline}
\label{method:MA-ROESL}
The hybrid three-phase training pipeline consists of three algorithmic components: 1) initial dataset collection, 2) rapid reward optimization, and 3) final policy fine-tuning. The corresponding pseudocode is depicted in Algorithm \ref{algo:MAROESL}.

\subsubsection{\textbf{Phase 1}}
\textbf{Initial Dataset Collection.}
In phase 1, the \texttt{VLM} is employed to automatically generate reward functions. The inputs consist of a video demonstration $v$ and the environment code $M$. Based on these inputs, the \texttt{VLM} generates a set of candidate reward functions $\mathcal{R}$. Each reward function $R_i \in \mathcal{R}$ is then used to train a policy $\pi_i$ in a simulation, and the resulting transition tuple $(s, a, R_i, s^{'})$ is collected. These transition tuples are stored in an offline dataset $\mathcal{D}$ for use in subsequent phases. To this end, all trajectories rollouted by trained policies are evaluated using the \texttt{VLM}, which in turn selects the best-performing reward function $R_{\text{sel}}$ along with its corresponding policy $\pi_{\text{sel}}$.

Although online training enables policy improvement through direct agent-environment interaction, it is time-consuming and computationally expensive. In practice, a reward function can only be evaluated after completing a full policy training cycle. When the reward fails to guide skill learning, this delayed feedback results in significant resource waste and poor training efficiency. To address this, the key challenge lies in accelerating the reward optimization process without compromising policy performance.

\subsubsection{\textbf{Phase 2}}
\textbf{Rapid Reward Optimization.} 
Offline RL significantly accelerates the training process by eliminating the need for agent-environment interaction~\cite{levine2020offline}, making it an effective solution for rapid reward optimization. Leveraging the offline dataset $\mathcal{D}$, which provides sufficient diversity to compare the relative effectiveness of VLM-generated reward functions, phase 2 efficiently evaluates and selects the most effective rewards. Specifically, each VLM-generated reward function $R_j^{'}$ is used to relabel the transition tuples from $\mathcal{D}$, replacing the original rewards to ensure consistency with the new reward functions. This results in updated transition tuples $(s, a, R_j^{'}, s^{'})$, where $R_j^{'}$ denotes the relabeled reward. Subsequently, a policy $\pi_j$ is trained using each relabeled reward function $R_j^{'}$, and the resulting behaviors are evaluated using the \texttt{VLM} to quantify the performance. By iteratively repeating this procedure, the reward function $R_{\text{bst}}$ corresponding to the best-performing behavior is rapidly identified, along with its associated policy $\pi_{\text{bst}}$.

Nevertheless, since offline RL exclusively learns from pre-collected behavior datasets, it inevitably suffers from distributional shift~\cite{fujimoto2019off}. This often leads the learned policies to exhibit unsafe behaviors, such as joint limit violations, when deployed on real robots. The key challenge thus becomes balancing efficient offline reward evaluation with the need for robust policy performance in real-world deployment.

\subsubsection{\textbf{Phase 3}}
\textbf{Final Policy Fine-tuning.} To mitigate the distribution shift inherent in offline policy learning, online fine-tuning has been shown to significantly enhance policy robustness and generalization~\cite{lee2022offline}. In phase 3, the policy $\pi_{\text{bst}}$, selected during offline evaluation, is used as the initial policy for online training, with the corresponding reward function $R_{\text{bst}}$ guiding the agent's interactions with the environment. As the agent collects new state-action pairs beyond the offline dataset $\mathcal{D}$, $\pi_{\text{bst}}$ adapts to previously unseen environment dynamics under the supervision of $R_{\text{bst}}$. Through this process, the fine-tuned policy $\pi_{\text{fin}}$ is obtained, achieving improved robustness and generalization when deployed in real-world scenarios.

\begin{algorithm}[t]
\caption{Hybrid three-phase training pipeline of MA-ROESL.}
\label{algo:MAROESL}
\begin{algorithmic}[1]
\small
\STATE \textbf{Require:} Environment code $M$, video demonstration $v$, vision-language model \texttt{VLM}, prompt $\texttt{prompt}$
\STATE \textbf{Hyperparameters:} Phase 1 iterations $N_1$, Phase 2 iterations $N_2$, Phase 1 reward functions $K_1$, Phase 2 reward functions $K_2$
\STATE \textbf{Initialize:} Offline dataset $\mathcal{D} \leftarrow \emptyset$

\STATE \texttt{// Phase 1: Initial Dataset Collection}
\FOR{$n = 1$ to $N_1$}
    \STATE $\mathcal{R_{\text{p}_\text{1}}} \leftarrow \texttt{VLM.GenerateRewards}(M, v, \texttt{prompt}, K_1)$
    \FOR{$i = 1$ to $K_1$}
        \STATE \texttt{// Access the $i$-th reward in $\mathcal{R_{\text{p}_\text{1}}}$}
        \STATE $(\pi_i, \mathcal{D}_i) \leftarrow \texttt{TrainAndCollect}(M, R_i)$
        \STATE $\mathcal{D} \leftarrow \mathcal{D} \cup \mathcal{D}_i$
        \STATE $\tau_i \leftarrow \texttt{Rollout}(M, \pi_i)$
    \ENDFOR
    \STATE $(R_{\text{sel}}, \pi_{\text{sel}}) \leftarrow \texttt{VLM.Evaluate}(v, \{\tau_i\}_{i=1}^{K_1})$
\ENDFOR

\STATE \texttt{// Phase 2: Rapid Reward Optimization}
\FOR{$n = 1$ to $N_2$}
    \STATE $\mathcal{R_{\text{p}_\text{2}}} \leftarrow \texttt{VLM.GenerateRewards}(M, v, \texttt{prompt}, K_2)$
    \FOR{$j = 1$ to $K_2$}
        \STATE \texttt{// Access the $j$-th reward in $\mathcal{R_{\text{p}_\text{2}}}$}
        \STATE $\mathcal{D}_j' \leftarrow \texttt{RelabelDataset}(\mathcal{D}, R_j)$
        \STATE $\pi_j \leftarrow \texttt{TrainOfflinePolicy}(M, \mathcal{D}_j')$
        \STATE $\tau_j \leftarrow \texttt{Rollout}(M, \pi_j)$
    \ENDFOR
    \STATE $(R_{\text{bst}}, \pi_{\text{bst}}) \leftarrow \texttt{VLM.Evaluate}(v, \{\tau_j\}_{j=1}^{K_2})$
\ENDFOR

\STATE \texttt{// Phase 3: Final Policy Fine-tuning}
\STATE $\pi_{\text{fin}} \leftarrow \texttt{OnlineFineTune}(M, \pi_{\text{bst}}, R_{\text{bst}})$

\STATE \textbf{Return:} Final fine-tuned policy $\pi_{\text{fin}}$
\end{algorithmic}
\end{algorithm}

\section{EXPERIMENTS}
In this section, extensive experiments were conducted to evaluate the capability of MA-ROESL to rapidly and efficiently learn skills from video demonstrations. The hardware setup is first introduced (Section~\ref{exp:hardware}), followed by experimental details (Section~\ref{exp:expsetup}). Subsequently, the skill learning performance of MA-ROESL was evaluated (Section~\ref{exp:skilllearning}), and the learned policy was deployed on the Unitree Go2 to assess MA-ROESL’s zero-shot sim-to-real transfer performance (Section~\ref{exp:sim2real}).

\subsection{Hardware Setup}
\label{exp:hardware}
The experiments were conducted using a Unitree Go2 quadruped robot. The detailed hardware configurations used for both deployment and training are summarized below:

\begin{itemize}
    \item \textit{On-board Computer:} NVIDIA Jetson Orin NX, offering 100\,TOPS of AI computing power, operating independently without external computational support.
    \item \textit{Training Platform:} NVIDIA IsaacGym simulator, running on a workstation equipped with an Intel i9-14900K CPU and a single NVIDIA RTX 4090 GPU.
\end{itemize}

\subsection{Experimental Setup}
\label{exp:expsetup}
The MA-ROESL framework was executed over four iterations with fixed hyperparameters. In phase 1 and phase 3, the online training algorithm was Proximal Policy Optimization (PPO)~\cite{schulman2017proximal}. In phase 2, rapid reward optimization was performed using Implicit Q-Learning (IQL)~\cite{kostrikov2022offline}. At each iteration, reward functions were generated and evaluated by GPT-4 Vision~\cite{2023GPT4VisionSC}.

The proposed method was evaluated using four video demonstrations of quadrupeds performing distinct skills: trot, pace, bound, and hop. These skills were selected to cover a range of typical quadrupedal locomotion skills, differing in coordination and dynamic characteristics. This diversity enabled the evaluation of the method’s capability to learn skills from varied video inputs and demonstrated its potential for scalable skill learning from diverse online video sources.

\subsection{Skill Learning Evaluation}
\label{exp:skilllearning}
\subsubsection{Training Efficiency Evaluation}
\begin{table}[!t]
\centering
\caption{Summary of Hyperparameters.}
\label{tab:hyperparameter}
\begin{tabular}{@{}l|ll@{}}
\toprule
\toprule
\text{Type} & \text{Parameter} & \text{Value} \\
\midrule
\multirow{6}{*}{\begin{tabular}[c]{@{}c@{}}Online Training\\ (PPO)\end{tabular}}
  & Number of Environments             & 4000   \\
  & Mini-batch Size                    & 2.4e4  \\
  & Discount Factor                    & 0.99   \\
  & GAE Discount Factor                & 0.95   \\
  & Entropy Regularization Coefficient & 0.01   \\
  & Learning Rate                      & 0.001  \\
\midrule
\multirow{5}{*}{\begin{tabular}[c]{@{}c@{}}Offline Training\\ (IQL)\end{tabular}}
  & Temperature                        & 3.0    \\
  & Target Network Update Coefficient  & 0.005  \\
  & Expectile                          & 0.7    \\
  & Discount Factor                    & 0.99   \\
  & Learning Rate                      & 3e-4   \\
\bottomrule
\bottomrule
\end{tabular}
\end{table}

\begin{table}[ht]
\caption{Training Efficiency Comparison of Different Locomotion Skills (in hours).}
\centering
\label{tab:time}
\begin{tabular}{lcccc}
\hline
\hline
\text{Locomotion Skills} & \text{SDS} & \textbf{MA-ROESL} & \text{Reduction(Percentage)}\\
\hline
\text{Trot} & 16.83 & \textbf{6.15} & 63.46\% \\
\text{Pace} & 12.20 & \textbf{3.64} & 70.16\% \\
\text{Bound} & 13.10 & \textbf{4.16} & 68.24\% \\
\text{Hop} & 15.83 & \textbf{4.21} & 73.40\% \\
\text{Average} & 14.49 & \textbf{4.54} & 68.67\% \\
\hline
\hline
\end{tabular}
\end{table}

\begin{figure}[ht]
    \centering
    \includegraphics[width=0.75\linewidth]{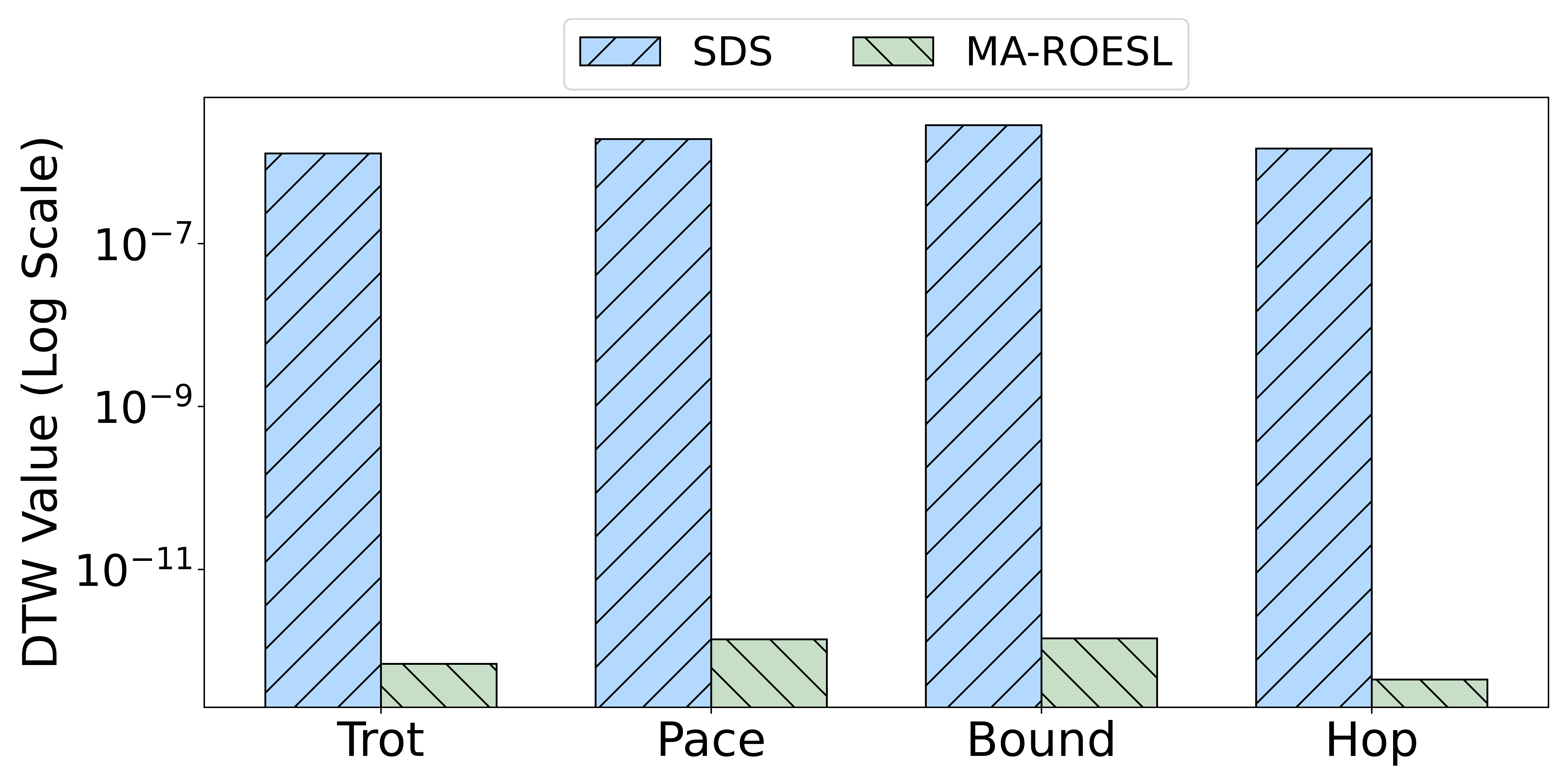}
    \caption{Evaluation of motion pattern alignment using dynamic time warping (DTW) analysis.}
    \label{fig:DTW}
\end{figure}

\begin{figure}[!t]
    \centering
    \includegraphics[width=0.9\linewidth]{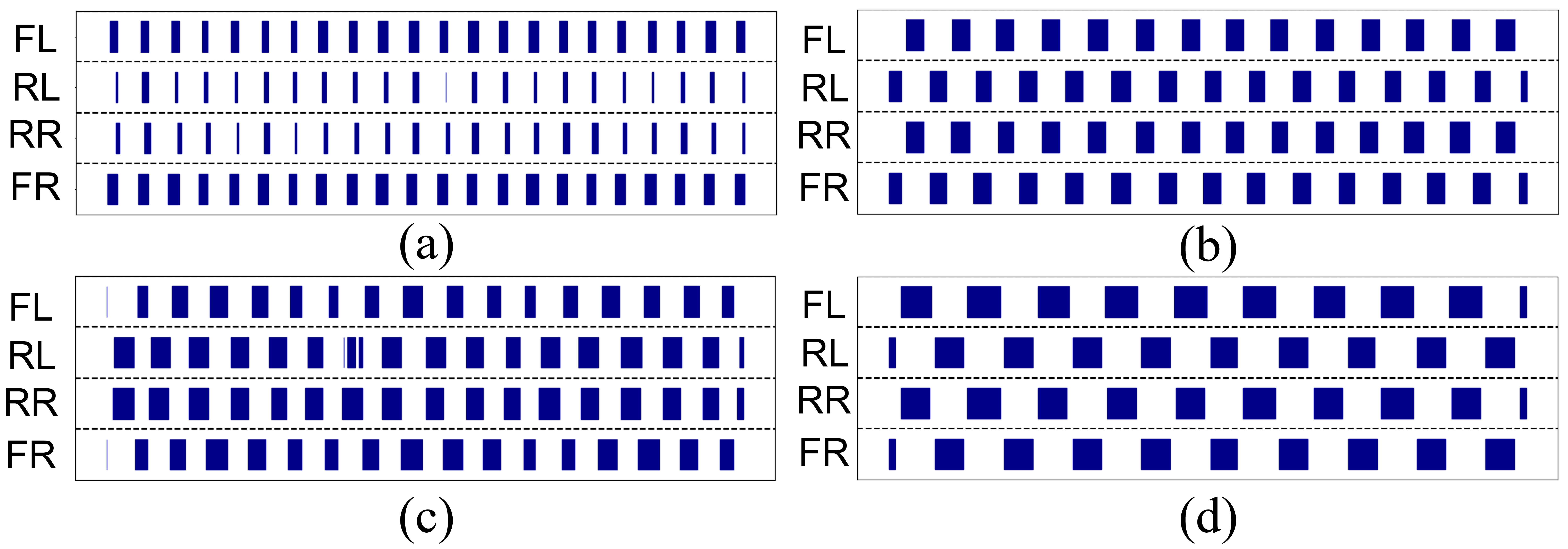}
    \caption{Contact patterns of the four limbs across different locomotion skills: a) Hop, b) Trot, c) Bound, and d) Pace, showing contact sequences of FL (Front Left), FR (Front Right), RL (Rear Left), and RR (Rear Right). Blue blocks indicate ground contact phases, while gaps denote swing phases.}
    \label{fig:contact_patterns}
\end{figure}

The training efficiency of MA-ROESL was assessed by comparing its per-skill training time with that of a state-of-the-art method, SDS~\cite{li2024sds}. The comparative efficiency evaluation across four skills is presented in Table~\ref{tab:time}. The relative reduction in training time was computed using the formula $\frac{t_{\text{SDS}} - t_{\text{MA-ROESL}}}{t_{\text{SDS}}} \times 100\%$, where $t_{\text{SDS}}$ and $t_{\text{MA-ROESL}}$ denote the per-skill training times achieved by SDS and MA-ROESL, respectively. The training time reductions highlight MA-ROESL’s efficiency in leveraging video demonstrations for rapid skill learning and its potential for scalable deployment in real-world robotic applications.

\subsubsection{Motion Patterns Alignment Evaluation}
\begin{figure*}[!t]
    \centering
    \includegraphics[width=\textwidth]{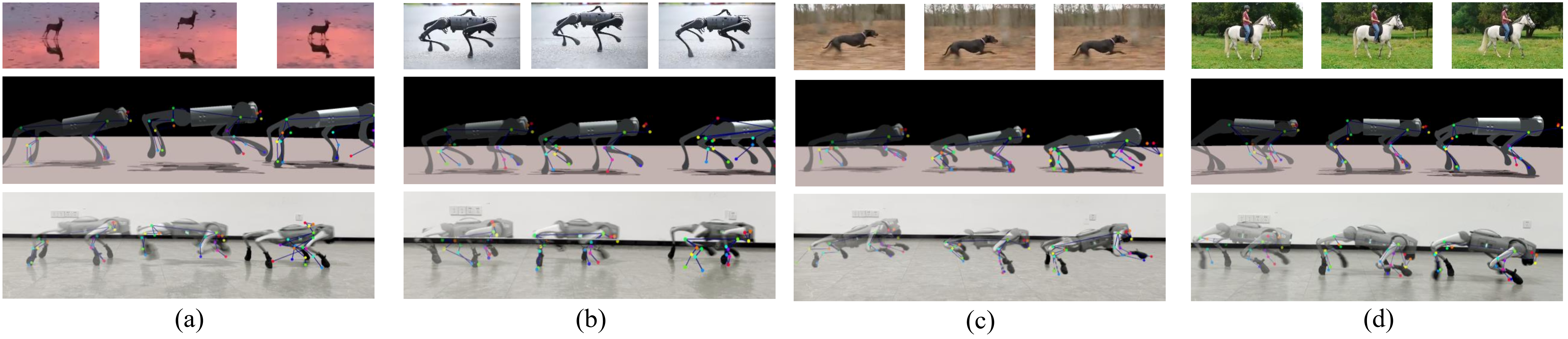}
    \caption{Simulation and real-world evaluations using MA-ROESL-trained policies for four locomotion skills: a) Hop, b) Trot, c) Bound, and d) Pace. The experimental video can be found in the supplementary materials.}
\label{fig:realworldresults}
\end{figure*}

\begin{figure}[ht]
    \centering
    \includegraphics[width=0.95\linewidth]{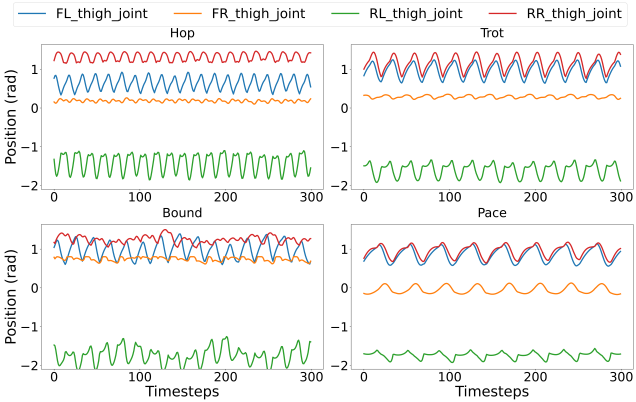}
    \caption{Illustration of thigh joint positions generated and recorded across a 300-step stable segment for four locomotion skills.}
    \label{fig:thigh_pos_evaluation}
\end{figure}
Similar to~\cite{li2024sds}, dynamic time warping (DTW) analysis was adopted to evaluate the alignment of motion patterns between demonstration videos and corresponding real-world deployment videos. Specifically, keypoint sequences were first extracted and serialized using a pose estimation method. These sequences were then utilized in the DTW evaluation. As shown in Fig.~\ref{fig:DTW}, MA-ROESL exhibits a lower DTW value across four skills, indicating that the skills learned by the agent were more closely aligned with the corresponding video demonstrations.

To further analyze whether the robot not only learned the demonstrated skills from videos but also captured their intrinsic dynamic characteristics, gait pattern analysis was conducted. As shown in Fig.~\ref{fig:contact_patterns}, the robot successfully reproduced diverse gait patterns consistent with video demonstrations. Moreover, thigh joint angles were recorded over stable segments comprising 300 steps to visualize cyclic motion profiles. Clear cyclic patterns were observed across most joints in Fig.~\ref{fig:thigh_pos_evaluation}. Additionally, an amplitude discrepancy between the rear left and front right thigh joints was identified. This asymmetry was attributed to the limitation of the input, which constrains the VLM’s ability to infer symmetric locomotion cues.

\subsection{Sim-to-Real Deployment Result}
\label{exp:sim2real}
To validate MA-ROESL's zero-shot sim-to-real performance, policies trained by MA-ROESL were deployed to Unitree Go2 without any sim-to-sim transfer. Fig.~\ref{fig:realworldresults} shows the trajectories of the robot performing different skills obtained from online videos, demonstrating reliable reproduction of the target skills in both simulated and real-world settings. This consistency demonstrated the effectiveness of MA-ROESL in zero-shot sim-to-real transfer.

\section{CONCLUSIONS}
In this work, we introduce MA-ROESL, a novel framework for efficient robot skill learning from video demonstrations. MA-ROESL incorporates a motion-aware frame selection method to enhance the fidelity of VLM inputs by capturing behaviorally salient frames, and employs a hybrid three-phase training pipeline that combines the offline training efficiency and online training adaptability. Simulation achieves a 68.67\% reduction in training time across four locomotion skills, reflecting a significant improvement in training efficiency. Real-world experiments validate the learned policies by reproducing locomotion skills closely aligned with the demonstrations. These results substantiate MA-ROESL’s robust zero-shot sim-to-real transfer capability and suggest its potential for scalable application. Building on this foundation, future work will explore learning skills from large-scale, weakly-labeled or unlabeled videos to further validate MA-ROESL's capability for skill learning in complex, long-horizon video demonstrations.











\bibliographystyle{IEEEtran}
\bibliography{root}

\end{document}